\documentclass[10pt,twocolumn,letterpaper]{article}

\usepackage[final]{cvpr}

\usepackage{amsmath,amssymb,amsfonts,amsthm}
\usepackage{graphicx}
\usepackage{booktabs}
\usepackage{multirow}
\usepackage{tabularx}
\PassOptionsToPackage{table}{xcolor}
\usepackage{colortbl}
\usepackage{enumitem}
\usepackage{subcaption}
\usepackage{pifont}

\usepackage{amsmath,amsfonts,bm}

\def\eqref#1{equation~\ref{#1}}

\def\1{\bm{1}}

\DeclareMathAlphabet{\mathsfit}{\encodingdefault}{\sfdefault}{m}{sl}
\SetMathAlphabet{\mathsfit}{bold}{\encodingdefault}{\sfdefault}{bx}{n}

\usepackage[most]{tcolorbox}
\newtcolorbox{findingsbox}{
  enhanced, breakable,
  colback=gray!6, colframe=gray!55,
  boxrule=0.4pt, arc=1.5pt,
  left=5pt, right=5pt, top=3pt, bottom=3pt,
  boxsep=1pt,
  fonttitle=\sffamily\bfseries\footnotesize,
  title={\textsc{Findings}},
  coltitle=black,
  attach title to upper={\quad}
}

\definecolor{cvprblue}{rgb}{0.21,0.49,0.74}
\usepackage[pagebackref,breaklinks,colorlinks,allcolors=cvprblue]{hyperref}

\def\confName{CVPR}
\def\confYear{2027}

\title{Query-Aware vs. Query-Agnostic Training-Free Video Token Pruning in Multimodal LLMs}

\author{Divya Kothandaraman, Timo Kumkel, Jaclyn Pytlarz \\
Dolby Laboratories, USA 
}

\begin{document}
\maketitle
\begin{abstract} 

Video Large Language Models process large numbers of visual tokens, making token pruning important for efficient inference. Training-free pruning methods differ in whether and how they use the user query, and the value of query conditioning remains unclear. We present a controlled study of training-free, pre-LLM video token selection. Using a shared Qwen-3-VL backbone, token budgets, and decoding settings, we compare 21 pruning principles spanning query-agnostic, shallow query-conditioned, structured query-conditioned, and holistic strategies. To instantiate the last two categories, we introduce Structured Query Pruning, which decomposes the question into semantic anchors, and a Pareto variant, which combines anchor relevance with query-agnostic visual diversity. Across Video-MME, MVBench, and LongVideoBench, relevance computed from the raw question embedding is unreliable (8--16 points below the best methods), whereas anchors derived from query decomposition avoid this failure. However, query access is not inherently advantageous: strong query-agnostic coverage methods remain competitive or superior in the full-field comparison. Relative to a redundancy-only baseline, structured selection helps most on evidence-specific and long-context tasks at tight budgets; on broad tasks, the holistic selector improves over either relevance or redundancy alone. Further, pruning reduces cost and memory. These results support a practical design rule: preserve visual coverage by default, and add structured relevance when the task and token budget demand selective evidence retrieval.

\end{abstract}
\section{Introduction}
Video-LLMs~\cite{yang2025qwen3,chen2024internvl} pair a visual encoder with an LLM to answer questions about video, and the sheer number of visual tokens is a primary bottleneck~\cite{xu2024survey,jin2025efficient}: self-attention scales quadratically with token count~\cite{feng2025breaking,shao2025holitom}; efficient variants such as FlashAttention~\cite{dao2022flashattention} reduce only the constant factor; and long videos inflate the KV cache and induce the ``attention sink'' phenomenon, which dilutes answer quality~\cite{luo2026quota,zhang2025dense}. This motivates high-ratio token pruning. \textit{Ante-hoc} methods filter tokens before the LLM, using frame-level similarity~\cite{huang2025prunevid}, cross-modal cosine alignment~\cite{cai2025flashvlm}, or learned selectors~\cite{zhang2025flexselect}. \textit{Intra-LLM} methods prune across layers using attention scores~\cite{ye2025atp,zhang2024sparsevlm} or KV-cache compression~\cite{wang2025adaretake,tao2025dycoke}. More recent work adds diversity-driven selection~\cite{alvar2025divprune}, spatiotemporal merging~\cite{shao2025holitom,fan2026flashvid}, and position-aware pruning~\cite{chien2025grounding,sun2026ivc}.

Despite this diversity, a basic question remains open: how much does knowing the query actually help, and under what conditions? We study \textbf{training-free, pre-LLM token pruning}, where visual tokens are selected before entering the language model. This isolates token selection from approaches that modify, merge, or prune representations inside the LLM, which are outside the scope of this study. Pre-LLM selection operates directly on video tokens, needs no training, and leaves the LLM unchanged; query-aware variants add at most one short text-only decomposition call.

\paragraph{Research questions.} Our analysis asks four questions: (RQ1)~\emph{When} does query information help, across budgets, task types, and video lengths? (RQ2)~Is \emph{shallow} conditioning sufficient, or is \emph{structured} decomposition necessary? (RQ3)~How should structured relevance be \emph{combined} with redundancy? (RQ4)~Does a query-derived relevance signal \emph{transfer} as a modular prior into other pruning algorithms? We address them in \autoref{sec:rq1}--\ref{sec:rq4}, respectively, with supporting analysis of the decomposition in \autoref{sec:anatomy} and efficiency in \autoref{sec:cost}.

\paragraph{Approach.} We organize methods along four levels of query use: query-agnostic selection, shallow query conditioning, structured query conditioning, and holistic selection that jointly optimizes relevance and redundancy (\autoref{sec:taxonomy}). To answer these questions on equal footing, we re-implement the core selection principle of a diverse set of published methods on Qwen3-VL under identical settings, so that methods differ only in which tokens they keep (\autoref{sec:testbed}). As the structured-query instance, we develop \textbf{Structured Query Pruning}, a training-free, plug-in method that decomposes the query ante-hoc into ``semantic anchors'' and a task policy to guide token scoring.

The results distinguish query access from effective query use. Direct alignment with the raw question can sharply reduce accuracy, whereas anchor-based relevance derived from query decomposition avoids this failure and produces a more stable relevance signal. Nevertheless, structured query conditioning does not dominate the full field: strong coverage- and position-aware selectors remain competitive or superior at many budgets. The central result is therefore not that query-aware pruning is always better, but that query information must be represented carefully and balanced against visual coverage. In brief: structured selection helps most at tight budgets on evidence-specific and long-context tasks (RQ1); raw-question similarity fails while decomposition-derived anchors do not (RQ2); holistic selection helps on broad tasks (RQ3); and the structured relevance prior provides selective gains when it complements the host method's existing selection signal (RQ4).

\noindent\textbf{Our primary contributions are:}
\begin{itemize}
\item \textbf{A controlled framework for studying query use in training-free, pre-LLM video token pruning.} We organize pruning strategies by how the query enters token selection and compare 21 selection principles, including our own, using identical visual features, token budgets, and decoding. This design isolates the selection criterion from differences in backbone and inference protocol.
\item \textbf{A systematic analysis of when query information helps.} Across three benchmarks, multiple retention budgets, a second backbone, and host-algorithm transfer experiments, we show that raw-query similarity is unreliable, that structured relevance helps most at tight budgets on evidence-specific and long-context tasks, and that strong query-agnostic coverage remains competitive.
\item \textbf{A structured-query selector and a Pareto relevance–coverage formulation.} Structured Query Pruning decomposes a question into semantic anchors and uses them to guide token selection. Its Pareto variant jointly preserves query relevance and visual diversity. Our analysis and ablations study when each component helps and where the method falls short.
\end{itemize}

\section{Background and Study Design}
\label{sec:design}
\label{sec:related}
A full survey of token pruning is beyond our scope (see~\cite{yao2026towards, shaosurvey}); we organize key works by the \emph{type of signal} used to judge token importance, following the four-level spectrum below. At each level we review prior work and then state how we re-implement or define its methods, choosing methods that cover the main signal types at that level; our structured-query instance sits at level~3 and our holistic selectors at level~4. We first describe the spectrum and the shared testbed on which all methods are compared.

\subsection{A Spectrum of Query Usage}
\label{sec:taxonomy}
We organize pruning strategies by how much query information is used and how it enters the selection decision, in four levels. \textbf{(1)~Query-agnostic} selection uses only intrinsic visual properties (magnitude, spatio-temporal redundancy, diversity), e.g., PruneVid~\cite{huang2025prunevid}, DyCoke~\cite{tao2025dycoke}, VisionZip~\cite{yang2025visionzip}, HoliTom~\cite{shao2025holitom}, DivPrune~\cite{alvar2025divprune}. \textbf{(2)~Shallow query conditioning} uses a direct low-level alignment signal, such as a cosine similarity between tokens and the raw question embedding, or an attention read-out inside the LLM: FlashVLM~\cite{cai2025flashvlm}, FastV-style attention~\cite{fastv2024,ye2025atp}, learned selectors~\cite{zhang2025flexselect}. \textbf{(3)~Structured query conditioning} first \emph{decomposes} the query into explicit entities, actions, attributes, and a task policy that drive token scoring; QuoTA~\cite{luo2026quota} (CoT query decouple) partially belongs here. \textbf{(4)~Holistic designs} combine structured relevance with query-agnostic redundancy and coverage through a joint objective.

\subsection{A Common Testbed and Experimental Setup}
\label{sec:testbed}
\paragraph{Re-implementations.}
To study the design space on equal footing, we implement the \emph{core selection principle} of 21 pruning methods, including the structured query and holistic methods, on the \emph{same} frozen Qwen3-VL visual tokens, so that methods differ \emph{only} in which video tokens they keep. We omit method-specific components that merge tokens, modify internal LLM states, or require training, to restrict the pruning at pre-LLM level. We define three shared per-token signals that many selectors draw on: magnitude $m_i=\lVert V_i\rVert_2$ ($S_{\text{init}}$), uniqueness $u_i$ (one minus the larger of the cosine similarity to the co-located token in the previous frame and the maximum cosine to its four spatial neighbours, zeroed above $\tau{=}0.98$; $S_{\text{red}}$), and anchor relevance $r_i$ ($S_{\text{sem}}$, zero under \textsc{Global} or without decomposition). Note that methods that use $r$ consume the structured-query prior. The four levels of \autoref{sec:taxonomy} are instantiated in \autoref{sec:level-agnostic}--\ref{sec:holistic}, each stating how its methods are re-implemented or defined.

\paragraph{Setup.}
All experiments use Qwen3-VL-8B-Instruct as a training-free plug-in, with cross-architecture validation on InternVL3-8B~\cite{chen2024internvl}. The budget is a percentage of visual tokens, and decoding is greedy with seed 42. Query decomposition is text-only and uses 7-shot in-context examples: we use Qwen3-VL-8B in text-only mode for the main comparisons (Secs.~\ref{sec:rq1}--\ref{sec:analysis-shallow}) and a Qwen3-0.6B decomposer for the controlled analyses, since a small pilot sweep showed no clear accuracy difference across decomposer sizes. We report accuracy on Video-MME (2700 QA pairs over 900 short/medium/long videos, no subtitles)~\cite{fu2025video}, MVBench (2996 videos)~\cite{li2024mvbench}, and LongVideoBench (1337 videos)~\cite{wu2024longvideobench}. Frame sampling, resolution, token-count rounding, InternVL3 settings, and dataset details are in \autoref{app:setup}.

\subsection{Level 1: Query-Agnostic Selection}
\label{sec:level-agnostic}
The largest family uses only intrinsic visual properties; the question never enters selection. \textit{Redundancy/similarity} methods drop overlapping tokens: PruneVid~\cite{huang2025prunevid} and Efficient Video Sampling~\cite{bagrov2025efficient} use cross-frame similarity, DART~\cite{wen2025stop} and VisionZip~\cite{yang2025visionzip} remove duplication, and KVTP~\cite{liu2025keyframe} isolates keyframes and prunes the rest. \textit{Diversity-driven} methods emphasize variance: DivPrune~\cite{alvar2025divprune} casts selection as a Max--Min Diversity Problem and Tango~\cite{yin2026tango} adds ST-RoPE locality priors (both host the structured-query prior in \autoref{sec:hosts}). InfoTok~\cite{ye2025infotok} proves via ELBO routing that uniform static budgeting is suboptimal. From this space we re-implement eight methods chosen to cover the main signal types (redundancy, diversity, position-aware, embedding-space): PruneVid, VisionZip, DyCoke~\cite{tao2025dycoke}, HoliTom~\cite{shao2025holitom}, TopV~\cite{yang2025topv}, IVC-Prune~\cite{sun2026ivc}, GAP~\cite{chien2025grounding}, and Representation Shift~\cite{choi2025representation} (the last six are reviewed under the families in \autoref{sec:other-families}). Each is reduced to its core selection rule on the shared magnitude $m$ and uniqueness $u$ signals, with merging, in-LLM, and training components omitted (details in \autoref{sec:cat-agnostic}).

\subsection{Level 2: Shallow Query Conditioning}
\label{sec:level-shallow}
These methods use the query only through a low-level alignment or attention signal: M-LLM~\cite{hu2025m} scores frames by text-conditioned importance, FlashVLM~\cite{cai2025flashvlm} applies cross-modal cosine filtering, FlexSelect~\cite{zhang2025flexselect} learns a selector mimicking cross-modal attention, and FRAME-VOYAGER~\cite{yu2024frame} retrieves frames from text. Inside the LLM, FastV~\cite{fastv2024, ye2025atp} and SparseVLM~\cite{zhang2024sparsevlm} prune with layer attention scores and token recycling. Such signals capture semantic proximity but not the logical structure of multi-step queries. We re-implement five representative of these signal types: FastV (raw query)~\cite{fastv2024}, FastV-Attn~\cite{ye2025atp}, FlashVLM, FlexSelect, and ReTaKe~\cite{wang2024retake} (reviewed in \autoref{sec:other-families}). Each becomes an ante-hoc scorer on the shared features, with signals inside the LLM replaced by proxies: raw-question cosine (FastV raw query, FlashVLM), a cross-modal attention read-out (FastV-Attn), the scoring rule of a learned selector (FlexSelect), or keyframe selection (ReTaKe) (details in \autoref{sec:cat-querybased}).

\subsection{Level 3: Structured Query Pruning}
\label{sec:methodology}
A smaller set of prior methods builds an explicit query representation before scoring. QuoTA~\cite{luo2026quota} decouples the query via CoT and has the LVLM rate frame-level relevance, with temporal chunking and bilinear merging; VIDEO-KTR~\cite{wang2026video} isolates Key Reasoning Tokens via reinforcement learning and counterfactual masking. Structured Query Pruning is our instance at this level (\autoref{sec:taxonomy}): a training-free, plug-in selector that scores visual tokens with query-derived semantic anchors, obtained by in-context decomposition into entity/action anchors and a task policy, before the LLM's generation phase. We introduce it as a concrete instance of structured query use, to probe the design space.

\subsubsection{Query Decomposition}

Decomposition is text-only: video frames are withheld to minimize overhead and avoid visual hallucination. Given the query, an LLM generates a structured JSON object by in-context learning containing a \emph{policy} (\textsc{Spatial}: static cues, localization, attributes; \textsc{Temporal}: motion and kinematic sequences; \textsc{Balanced}: both; \textsc{Global}: coarse semantic understanding or summarization), \emph{entities} and \emph{actions} (bare-noun and bare-verb anchors, e.g., ``socks'', ``jumping''), and their \emph{correlates} (disambiguating attributes or modifiers, e.g., socks: [red]). Anchors are derived from the query text, and generic meta-verbs such as ``showing'' are prohibited. 

\subsubsection{Policy-Driven Importance Scoring}
\label{sec:importance_scoring}
Intrinsic saliency $S_{\text{init}}$ is the per-token $L_2$ norm of the visual embeddings, normalized per sample, so that visually dense regions are preserved regardless of the query. The semantic map $S_{\text{sem}}$ aligns tokens with anchors: for each anchor we take the cosine similarity between every visual token and its text embedding (negatives clamped to zero), add a correlate \emph{boost} (the mean similarity to the anchor's correlate embeddings), calibrate bare anchors with the boost-to-similarity ratio of attributed anchors so they are not overshadowed, and sum over anchors, $S_{\text{sem}} = \sum (\text{similarity}_{\text{anchor}} + \text{boost}_{\text{correlate}})$. The final score is $S = \mathrm{norm}\!\left(S_{\text{init}} + S_{\text{sem}}\right)$, re-normalized per sample; under the \texttt{GLOBAL} policy, $S_{\text{sem}}\!\equiv\!0$ and $S$ reduces to $S_{\text{init}}$. Keeping the top-$k$ tokens by $S$ is the \textsc{Structured} selector.

More details and a diagrammatic representation are in App. \autoref{fig:overview} and App.~\ref{app:method}.

\subsubsection{Host Algorithms}
\label{sec:hosts}
The structured query prior is also designed to be modular. We inject the anchor relevance $r$ into five host algorithms: QuoTA~\cite{luo2026quota}, AdaReTaKe~\cite{wang2025adaretake}, DivPrune~\cite{alvar2025divprune}, Tango~\cite{yin2026tango}, and SemanticEvidence~\cite{li2026keeping}. Each host keeps its own selection philosophy and receives $r$ in its native ranking criterion: QuoTA mixes $r$ into its token-level quality score under an adaptive threshold with a per-frame coverage guarantee; AdaReTaKe starts from a blend of the fused score and uniqueness and refines under-covered frames and cells; DivPrune uses the fused score as the value term of its greedy Max--Min diversity selection; Tango smooths the fused importance over time, adds a temporal-consistency term, and selects with segment-balanced quotas; and SemanticEvidence accumulates evidence from $r$ and uniqueness along per-position temporal trajectories and keeps the strongest evidence chains. 
Details in \autoref{app:c-hosts}.

\subsection{Level 4: Holistic Selection}
\label{sec:holistic}
Holistic selectors combine structured relevance with query-agnostic redundancy and coverage through a joint objective (\autoref{sec:taxonomy}). We define two: \textsc{Nested}, a sequential variant, and \textsc{Pareto}, a joint multi-objective selector (details in \autoref{sec:cat-ours}).

\subsubsection{Nested Selection}
\textsc{Nested} filters tokens in two stages: it first keeps a $2k$ pool of the most unique tokens by uniqueness $u$ (the PruneVid criterion), then re-ranks that pool by the fused relevance $S$ and keeps the top $k$.

\subsubsection{Hybrid Pareto Selection}
Redundancy-based pruning (e.g., PruneVid) removes visual overlap but ignores the user's semantic goals; query-based pruning ensures relevance but may retain redundant tokens that share an anchor. We therefore propose a hybrid, based on multi-objective optimization, for the \texttt{SPATIAL}, \texttt{TEMPORAL}, and \texttt{BALANCED} policies (\texttt{GLOBAL} defaults to redundancy-based sampling); it is our \emph{holistic} design in the spectrum. 

We experimented with simple heuristic score combinations---additive fusion ($S_{\text{red}} + S_{\text{sem}}$), multiplicative fusion ($S_{\text{red}} \cdot S_{\text{sem}}$), and weighted linear combinations---but these consistently underperformed both Nested and Pareto, because additive and multiplicative combinations collapse two distinct objective dimensions into a single scalar, losing the ability to preserve tokens that are excellent along one dimension even if mediocre along the other. This motivates our \textbf{Pareto} multi-objective formulation.

We maximize two potentially conflicting objectives, the diversity score $S_{\text{red}}$ (normalized PruneVid uniqueness) and the fused relevance $S$, by iteratively extracting Pareto frontiers (scanning tokens in descending $S_{\text{red}}$, a token joins the frontier if its $S$ strictly exceeds the running maximum). If a frontier fits within the remaining budget it is accepted whole and removed from the pool; otherwise its tokens are ranked by the product $P(i)=S_{\text{red},i}\cdot S_i$ and the highest are kept until the budget is met, and the same product fills any residual budget. 

Full details of the hybrid Pareto selector are in \autoref{app:method}.

\subsection{Other Families of Pruning Methods}
\label{sec:other-families}
\label{sec:related-intrallm}
These methods fall outside our four-level taxonomy or operate within the LLM. We include them here to provide a complete view of the token-reduction landscape. Where applicable, we evaluate their core selection principles within our controlled pre-LLM setting; methods whose defining mechanisms cannot be meaningfully instantiated before the LLM are discussed for context but excluded from the empirical comparison.

\paragraph{Preserving continuity.} GAP~\cite{chien2025grounding} freezes the coordinate grid and IVC-PRUNE~\cite{sun2026ivc} protects implicit visual coordinates in RoPE against scrambled positions; Video-XL-Pro~\cite{liu2025video} and GRT~\cite{zhang2025dense} handle motion via reconstructive compression and motion-gated inter-tokenizers. These priors are largely orthogonal to the query-usage spectrum.

\paragraph{Intra-LLM algorithms \& cache optimization.} AdaptInfer~\cite{zhang2025adaptinfer} (text-prior re-ranking), AdaRETAKE~\cite{wang2025adaretake} and RETAKE~\cite{wang2024retake} (adaptive KV budgets, keyframe isolation), DyCoke~\cite{tao2025dycoke} (decoding-time pruning and merging), LearnPruner~\cite{takezoe2026learnpruner} (learnable early-layer scorer), TopV~\cite{yang2025topv} (Sinkhorn transport), Representation Shift~\cite{choi2025representation} (embedding-shift scores), and VideoNSA~\cite{song2025videonsa} (native sparse attention for 128K contexts) prune across layers or the KV cache.

\paragraph{Advanced inference paradigms.} These include speculative decoding (SpecVLM~\cite{huang2025specvlm}, verifier-guided tree pruning), streaming (rLiVS~\cite{dorovatas2025recurrent}, a FIFO memory queue; CacheFlow~\cite{patel2025cacheflow}, GPU--CPU KV streaming), autoregressive token selection (AutoGaze~\cite{shi2026attend}), and joint generation and classification (BiGain~\cite{liu2026bigain}, spectral tokens).

\paragraph{Token merging.} Merging consolidates rather than discards tokens: HoliTom~\cite{shao2025holitom}, FlashVID~\cite{fan2026flashvid}, and HieraVid~\cite{guo2026hieravid} merge hierarchically over space and time; ToMe~\cite{bolya2022token} uses bipartite matching, DTEM~\cite{lee2024learning} a learned grouping embedding, and Video Token Merging~\cite{pollard2025video} temporal merging across adjacent frames.

\section{Results}
\label{sec:results-main}
\begin{figure*}[!t]
    \centering
    \includegraphics[width=\textwidth]{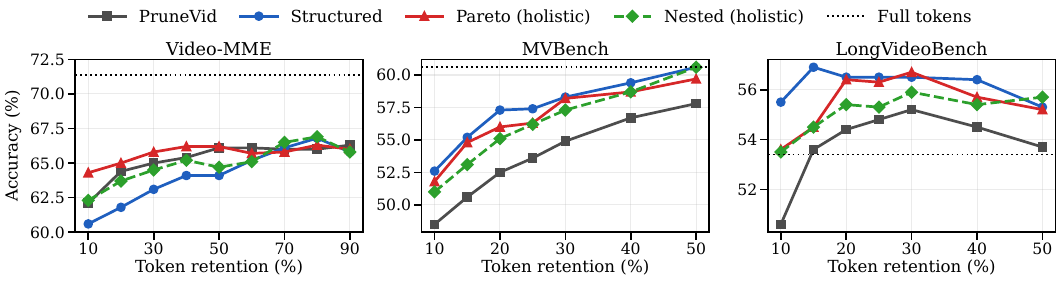}
    \caption{\textbf{Accuracy vs.\ token retention} on Video-MME, MVBench, and LongVideoBench (Qwen3-VL-8B). Dotted line: full-token baseline (71.4, 60.6, 53.4). Numerical tables are in \autoref{app:results}.}
    \label{fig:budget}
\end{figure*}

We answer the four research questions in turn (RQ1--RQ4), then analyze the query decomposition itself and the efficiency of pruned inference. 

\subsection{RQ1: When Does Query Information Help?}
\label{sec:rq1}
To study how the query enters selection, we compare the \textbf{Redundancy-Aware} baseline (PruneVid uniqueness, query-agnostic), the \textbf{Structured-Query} selector, and two holistic selectors: \textbf{Nested} (PruneVid filtering to a $2K$ pool, then re-ranking by $S$) and \textbf{Pareto} (joint multi-objective). 
\paragraph{Structure helps on evidence-specific and long-context tasks, a balanced approach helps on broader tasks.} As shown in \autoref{fig:budget}, Video-MME highlights the importance of balancing semantic relevance with visual coverage, as structured-query pruning alone is below redundancy-aware selection at every budget from 10\% to 60\%. Pareto provides this balance and achieves the strongest results among the four selectors. Against PruneVid, structured-query selection gains 2--5 points on MVBench (4.1 at 10\% and 4.8 at 20\%) and up to 4.9 points on LongVideoBench at 10\% retention, where critical evidence appears only briefly amid large amounts of irrelevant content. The edge is largest at the tightest budget and erodes as the budget grows ($+4.9\!\to\!+1.3$ on LongVideoBench from 10\% to 30\%; $+4.8\!\to\!+2.8$ on MVBench from 20\% to 50\%). We hypothesize that query structure helps most when the budget is too small for even good coverage to preserve the relevant evidence. 

\paragraph{The evidence filtering effect.} Several compressed configurations \emph{match or exceed} the full-token baseline. On MVBench at 50\% retention, structured-query pruning ties the 60.6\% baseline; at 40\% retention on LongVideoBench it reaches 56.4\% versus the 53.4\% full-token result. A plausible explanation is \emph{evidence filtering}: removing irrelevant content may raise the signal-to-noise ratio and reduce attention dilution. Token selection can thus match or exceed full-token accuracy in some settings. On LongVideoBench the effect is not specific to query use: even the query-agnostic PruneVid exceeds the full-token baseline from 15\% retention (53.6--55.2\%), and structured selection adds up to $+4.9$ points on top.

\paragraph{Task dependence.} Overall, the best strategy depends on the task: redundancy-aware methods determine which information is \emph{unique}, structured-query methods which is \emph{useful}, and combining them helps on Video-MME, whereas on LongVideoBench and MVBench, pure structured selection is the strongest of the four selectors (App.~\ref{app:e-synth}).

\paragraph{The trends hold on a second backbone.} As a check that the trends are not specific to Qwen3-VL, we evaluate on InternVL3-8B~\cite{chen2024internvl}, which uses a different visual encoder and projection than Qwen3-VL. Table~\ref{tab:longvideo_internvl} reports LongVideoBench results, a particularly challenging setting for pruning. Structured-query selection outperforms PruneVid by 0.7--1.8 points across budgets on InternVL3, with the holistic Pareto selector strongest at moderate compression (20--30\%). At 10\% budget the structured query method reaches 48.7\% versus 46.9\% for PruneVid (+1.8).

\begin{table}[t]
\centering
\caption{LongVideoBench accuracy (\%) on InternVL3-8B.}
\label{tab:longvideo_internvl}
\resizebox{\columnwidth}{!}{ 
\begin{tabular}{c|c|c|>{\columncolor{gray!12}}c>{\columncolor{gray!12}}c
}
\toprule

& \textbf{Redundancy}
& \textbf{Structured query}
& \multicolumn{2}{c}{\textbf{Holistic}} \\

\textbf{Budget (\%)}
& \textbf{PruneVid}
& \textbf{Structured} & \textbf{Pareto} \\

\midrule

10 & 46.9 & \textbf{48.7} & 48.7 \\
20 & 49.3 & 50.5 & \textbf{51.0} \\
30 & 50.0 & 51.4 & \textbf{51.6}  \\
40 & 50.1 & \textbf{51.3} & 50.5  \\
50 & 51.1 &  \textbf{51.8} & 51.5 \\
\midrule
100 (Upper bound) & \multicolumn{4}{c}{51.8} \\

\bottomrule
\end{tabular}
}
\end{table}

\begin{findingsbox}
Query structure pays off most when the budget is tight and the evidence is sparse.
\end{findingsbox}

\subsection{RQ2: How Should the Query Be Used? Shallow vs.\ Structured}
\label{sec:analysis-shallow}
\label{sec:analysis-crossbench-sota}
We run all baselines on MVBench ($N{=}2996$) and LongVideoBench ($N{=}1337$) at the two most aggressive budgets (10\% and 20\% retention), and Video MME at 20\% retention using the shared-testbed re-implementations of \autoref{sec:testbed} and Qwen3-VL-8B. \autoref{tab:fullfield}, \autoref{tab:sota_videomme} are organized by the four levels of \autoref{sec:taxonomy} and separates \textbf{external baselines}, every method whose selection does \emph{not} use the structured query prior (levels 1 and 2: the query-agnostic methods, and the query-based own-signal methods), from \textbf{methods using the structured query prior} (levels 3 and 4): the structured query selector, the five host algorithms into which we inject $r$ ($\dagger$) and the holistic selectors. For the host algorithms, the variants apply the structured relevance prior (r) within existing host algorithms. 

\begin{table}[t]
\centering
\footnotesize
\setlength{\tabcolsep}{3.2pt}
\caption{\textbf{The full field on LongVideoBench ($N{=}1337$) and MVBench ($N{=}2996$)}: accuracy (\%) at 10\% and 20\% token retention, Qwen3-VL model and Qwen3-VL-8B decomposer. $^\dagger$: uses our structured-query prior $r$; 
Best per column in \textbf{bold}.}
\label{tab:fullfield}
\begin{minipage}[t]{1.0\columnwidth}

\centering
\begin{tabular}[t]{lcccc}
\toprule
& \multicolumn{2}{c}{LongVideoBench} & \multicolumn{2}{c}{MVBench} \\
\cmidrule(lr){2-3}\cmidrule(lr){4-5}
Method & 10\% & 20\% & 10\% & 20\% \\
\midrule
\multicolumn{5}{l}{\emph{Level 1: Query-agnostic (external baselines)}} \\
IVC-Prune & \textbf{57.1} & 56.0 & 52.7 & 58.0 \\
GAP & 56.8 & 56.8 & 53.8 & 59.1 \\
Representation-Shift & 56.7 & 56.8 & 54.1 & 59.0 \\
VisionZip & 55.8 & \textbf{57.1} & 53.1 & 58.3 \\
PruneVid & 50.6 & 54.4 & 48.5 & 52.5 \\
DyCoke & 51.2 & 54.4 & 44.9 & 54.4 \\
HoliTom & 50.6 & 53.7 & 48.4 & 53.0 \\
TopV & 46.0 & 47.1 & 45.4 & 50.6 \\
\midrule
\multicolumn{5}{l}{\emph{Level 2: Query-based, own signal (external baselines)}} \\
FastV (raw-query) & 45.6 & 48.8 & 39.4 & 43.9 \\
FastV-Attn & 57.0 & 56.8 & 53.8 & 59.0 \\
FlashVLM & 45.8 & 49.0 & 39.3 & 43.7 \\
FlexSelect & 55.4 & 56.3 & 53.9 & \textbf{59.5} \\
ReTaKe & 48.0 & 52.0 & 40.3 & 46.9 \\
\midrule
\multicolumn{5}{l}{\emph{Level 3: Structured query and hosts ($^\dagger$ = uses prior $r$)}} \\
Structured$^\dagger$ & 55.5 & 56.5 & 52.6 & 57.3 \\
QuoTA$^\dagger$ & 56.6 & 56.9 & \textbf{54.9} & 58.4 \\
AdaReTaKe$^\dagger$ & 53.4 & 56.2 & 52.2 & 56.8 \\
DivPrune$^\dagger$ & 55.9 & \textbf{57.1} & 54.6 & 57.9 \\
Tango$^\dagger$ & 54.5 & 56.8 & 52.5 & 56.4 \\
SemEv$^\dagger$ & 50.9 & 53.2 & 46.9 & 53.0 \\
\midrule
\multicolumn{5}{l}{\emph{Level 4: Holistic}} \\
Pareto$^\dagger$ & 53.6 & 56.4 & 51.8 & 56.0 \\
Nested$^\dagger$ & 53.5 & 55.4 & 51.0 & 55.1 \\
\bottomrule
\end{tabular}
\end{minipage}
\end{table}

\paragraph{How the query enters selection is more important than query access alone.} The raw-question proxies, including the raw-query FastV variant and FlashVLM-style cosine filtering, trail the best shared-testbed methods by 8–16 points on MVBench and LongVideoBench. This does not imply that all shallow conditioning fails: attention-style FastV and FlexSelect perform substantially better. Instead, the result shows that direct similarity to a pooled question embedding is an unreliable pruning signal. Structured decomposition avoids this collapse. Note that this contrast is between a single pooled question embedding and word-level anchors produced by LLM decomposition. The anchors in the structured query method can be viewed as keywords, refined by decomposition: keywords taken directly from a query are often messy (generic or meta words, entities and actions mixed together), whereas decomposition keeps bare entities and actions, adds disambiguating attributes only when needed, and drops generic verbs.

\paragraph{Against the full field, structured query usage is competitive but not dominant.} As shown in \autoref{tab:sota_videomme}, on Video-MME at 80\% pruning, the holistic \textsc{Pareto} method (65.0\%) clearly beats shallow conditioning (FlashVLM 57.9, FlexSelect 60.8, FastV 61.7) and position-aware or weak query-agnostic methods (IVC-Prune 61.2, GAP 61.5, TopV-Sinkhorn 54.5). But the strongest query-agnostic coverage baselines (DyCoke 66.9, VisionZip 65.8) exceed it. Thus, the full-field comparison shows that structured query conditioning is competitive, but not dominant. On both MVBench and LongVideoBench, several query-agnostic baselines perform at or near the top of the field. Structured selection consistently outperforms the shallow query-conditioned baselines, but remains mid-field relative to the strongest query-agnostic methods. Incorporating the structured relevance prior into existing host methods (\autoref{tab:hosts}) helps in certain cases. Thus, structured query relevance can complement strong visual selection in specific scenarios.

\begin{table}[t]
\centering
\caption{Comparison with recent token pruning principles on Video-MME at 80\% pruning ratio (20\% retention), Qwen3-VL backbone; higher is better, best in \textbf{bold}.}
\label{tab:sota_videomme}
\resizebox{\columnwidth}{!}{ 
\begin{tabular}{llc}
\toprule
Method & Type & Acc.\,(\%) \\
\midrule
\multicolumn{3}{l}{\emph{Level 1: Query-agnostic}} \\
TopV-Sinkhorn~\cite{yang2025topv} & Transport & 54.5 \\
IVC-Prune~\cite{sun2026ivc} & Position-aware & 61.2 \\
GAP-FeatureNorm~\cite{chien2025grounding} & Grounding-aware & 61.5 \\
Representation Shift~\cite{choi2025representation} & Embedding-space & 61.6 \\
PruneVid~\cite{huang2025prunevid} & Redundancy & 64.4 \\
HoliTom~\cite{shao2025holitom} & Redundancy & 64.5 \\
VisionZip~\cite{yang2025visionzip} & Redundancy & 65.8 \\
DyCoke~\cite{tao2025dycoke} & Redundancy & \textbf{66.9} \\
\midrule
\multicolumn{3}{l}{\emph{Level 2: Shallow query conditioning}} \\
FlashVLM~\cite{cai2025flashvlm} & Shallow query & 57.9 \\
RETAKE-DPSelect~\cite{wang2024retake} & Keyframe selection & 60.0 \\
FlexSelect~\cite{zhang2025flexselect} & Learned selector & 60.8 \\
FastV~\cite{ye2025atp} & Shallow (attention) & 61.7 \\
\midrule
\multicolumn{3}{l}{\emph{Level 3: Structured query}} \\
Structured & Structured query & 61.8 \\
\midrule
\multicolumn{3}{l}{\emph{Level 4: Holistic}} \\
Nested & Sequential holistic & 63.7 \\
Pareto & Structured + holistic & 65.0 \\
\bottomrule
\end{tabular}
}
\end{table}

\begin{findingsbox}
How the query enters selection matters more than whether it does; decomposed anchors succeed where raw-question similarity fails.
\end{findingsbox}

\subsection{RQ3: How Should Relevance Be Combined with Redundancy?}
\label{sec:rq3}

The preceding results already reveal an interaction between relevance and redundancy across tasks and token budgets. We therefore analyze this relationship more deeply. We compare \textsc{Nested}, which first filters tokens by redundancy and then re-ranks the surviving pool by structured relevance, with \textsc{Pareto}, which jointly optimizes redundancy-based diversity and structured relevance.

Joint optimization is most useful when neither objective alone is sufficient. On Video-MME, \textsc{Pareto} is strongest across the aggressive 10--50\% retention range, reaching 64.3\% at 10\% retention compared with 62.1\% for redundancy-only selection and 60.6\% for structured selection. It also consistently exceeds the sequential \textsc{Nested} variant in this regime. This suggests that early hard filtering can discard tokens that are weak under one criterion but valuable under the other, whereas Pareto selection preserves tokens that are strong along either dimension. 

This benefit is not universal. On MVBench and LongVideoBench, structured relevance alone is generally as good as or better than either holistic selector, particularly at the tightest budgets. Thus, relevance and redundancy should not be combined indiscriminately: when broad visual coverage and query-specific evidence are both important, treating them as separate objectives is preferable to sequential filtering; when the task is dominated by sparse, query-specific evidence, adding a redundancy objective can dilute the stronger relevance signal. Overall, our results favor \textsc{Pareto} over \textsc{Nested} when hybridization is warranted, while showing that the decision to hybridize should itself depend on the task and compression regime.




\begin{findingsbox}
Hybridize when both coverage and query evidence matter, and prefer joint optimization over sequential filtering.
\end{findingsbox}

\subsection{RQ4: Does the Relevance Prior Transfer to Other Methods?}
\label{sec:rq4}
We inject the anchor relevance $r$ into five host algorithms (\autoref{sec:hosts}) and compare each host with and without the prior.

\begin{table}[t]
\centering
\footnotesize
\setlength{\tabcolsep}{3.5pt}
\caption{\textbf{Is the structured-query prior transferable?} Video-MME accuracy (\%), Qwen3-VL-8B with a Qwen3-0.6B decomposer. Each host with and without our prior at 20\% retention. }
\label{tab:hosts}
\begin{tabular}[t]{lrrr}
\toprule
Host & w/o $r$ & w/ $r$ & $\Delta$ \\
\midrule
QuoTA & 61.41 & 62.00 & $+0.59$ \\
AdaReTaKe & 65.81 & 64.04 & $-1.78$ \\
DivPrune & 63.74 & 64.19 & $+0.44$ \\
Tango & 62.00 & 62.78 & $+0.78$ \\
SemanticEvidence & 64.37 & 63.67 & $-0.70$ \\
\bottomrule
\end{tabular}
\end{table}

\paragraph{The benefits of the structured relevance prior depend on the host method.} Query relevance can provide complementary information beyond existing coverage signals. We integrate the anchor relevance into each host's native ranking criterion while preserving its selection philosophy; the protocol is in \autoref{app:c-hosts}. Injecting $r$ into five host algorithms at 20\% retention (\autoref{tab:hosts}) changes accuracy by $+0.78$ (Tango), $+0.59$ (QuoTA), and $+0.44$ (DivPrune), and by $-1.78$ (AdaReTaKe) and $-0.70$ (SemanticEvidence). On MVBench and LongVideoBench too, the prior helps in some cases. Since structured relevance identifies \emph{what} is relevant while diversity and coverage mechanisms determine \emph{how broadly} it is represented, these results suggest that strong coverage methods already capture much of what structured relevance provides when the token budget is sufficient. We hypothesize that its value increases under more aggressive compression, where retaining broad visual coverage becomes difficult and the selector must instead identify the small subset of evidence most relevant to the query. 

\begin{findingsbox}
The prior helps selectively, mainly where a host’s own signal lacks query relevance.
\end{findingsbox}

\subsection{Further Analysis: Anatomy of the Query Decomposition}
\label{sec:anatomy}
\paragraph{Policy routing and spatio-temporal smoothing.}
Different policies adapt the utilization of semantic signals: \texttt{GLOBAL} defaults to redundancy-based selection, \texttt{BALANCED} uses only primary anchors without correlates, and \texttt{SPATIAL}/\texttt{TEMPORAL} leverage the full anchor-correlate hierarchy. To prevent fragmented selection from localized semantic peaks, we apply policy-dependent spatio-temporal Gaussian smoothing over the importance map reshaped as $[T,H,W]$: spatial tasks emphasize spatial density $(w_s{=}0.75, w_t{=}0.25)$, temporal tasks prioritize motion continuity $(w_s{=}0.25, w_t{=}0.75)$, and balanced queries use equal weighting.

Table~\ref{tab:context_ablation} shows four observations. First, explicit query analysis provides the largest gain: removing query decomposition and semantic anchoring entirely drops accuracy from 62.8\% to 56.5\%, so within our structured query method, query decomposition is the component that contributes most under aggressive compression. Second, policy-specific processing shows a small difference within this component ablation: moving from \texttt{BalancedSmooth} to \texttt{PolicySmooth, NoCorr} improves 61.3\%$\to$61.9\%. Third, correlate-based context boosting changes accuracy in specific cases (\texttt{PolicySmooth, NoCorr} $\to$ \texttt{PolicySmooth}: 61.9\%$\to$62.0\%). Fourth, intrinsic visual saliency initialization gives the largest change among scoring refinements (62.0\%$\to$62.8\%), suggesting that semantic relevance and visual saliency are complementary. Taken together, the results identify query-derived anchors as the primary source of improvement, with policy routing and auxiliary correlates contributing more selectively.

\begin{table}[t]
\centering
\caption{\textbf{Ablation on Video-MME} (fixed 20\% budget, Qwen3-VL-8B decomposer). Each row progressively enables structured-query components.}
\label{tab:context_ablation}
\resizebox{0.7\columnwidth}{!}{ 
\begin{tabular}{lc}
\toprule
Variant & Accuracy (\%) \\
\midrule
NoSmooth, MagInit, NoCorr, NoDecomp & 56.5 \\
BalancedSmooth & 61.3 \\
PolicySmooth, NoCorr & 61.9 \\
PolicySmooth & 62.0 \\
PolicySmooth, MagInit & \textbf{62.8} \\
\bottomrule
\end{tabular}
}
\end{table}

\paragraph{Policy Distribution.} Table~\ref{tab:policy_dist} reports the model assigned policies. MVBench has the most TEMPORAL queries (70.5\%), consistent with its focus on action recognition and temporal ordering; Video-MME is more balanced, with the most GLOBAL queries (28.7\%), reflecting its diverse question types; and LongVideoBench is dominated by TEMPORAL (46.8\%) and SPATIAL (38.8\%) queries, reflecting the need for precise localization over long sequences.

\begin{table}[t]
\centering
\caption{Distribution of query policies across benchmarks (extracted from the decomposition cache). MVBench counts are over the 2706 successfully parsed samples.}
\label{tab:policy_dist}
\resizebox{\columnwidth}{!}{
\begin{tabular}{lcccc}
\toprule
Benchmark & SPATIAL & TEMPORAL & BALANCED & GLOBAL \\
\midrule
Video-MME ($n$=2700) & 26.1\% & 31.9\% & 13.2\% & 28.7\% \\
MVBench ($n$=2706) & 12.3\% & 70.5\% & 14.0\% & 3.1\% \\
LongVideoBench ($n$=1337) & 38.8\% & 46.8\% & 12.2\% & 2.2\% \\
\bottomrule
\end{tabular}
}
\end{table}

\subsubsection{Decomposition-Overhead Reduction Study}
\label{sec:addl-cot}

\paragraph{Protocol.}
Using \textsc{Pareto} selection on $100$ Video-MME samples at a $20\%$ budget, we
sweep (A) the decomposer size (Qwen3-0.6B / 1.7B / 4B and the 8B VLM-as-decomposer), (B)
batched-decomposition per-query latency at batch sizes $\{1,4,8,16\}$, and (C) the decomposition
\texttt{max\_new\_tokens} $\in\{512,256,128\}$.

\paragraph{Results.} Accuracy varies little across decomposer sizes, while batching and shorter generation substantially cut per-query latency.


\subsection{Efficiency}
\label{sec:cost}
Pruning to 20\% retention cuts prefill compute by $4.56\times$ ($1207.8\to264.6$ TFLOPs) and peak memory from $40.7$ to $27.0$\,GB. Even after paying the shared $3.06$\,s decomposition cost, \textsc{Structured} runs end to end in $17.3$\,s versus $21.4$\,s without pruning, because the savings from $80\%$ fewer tokens more than offset the decomposition overhead. The decomposition itself is cheap: accuracy varies little across decomposer sizes, so we observe no clear benefit from the larger decomposer, while the $0.6$B model runs $1.7\times$ faster per query; batching reduces per-query latency from $3.23$\,s to $0.69$\,s (bs$=16$), and shortening generation to $128$ tokens cuts it to $1.82$\,s with no observable accuracy change. Query-agnostic methods avoid the decomposition step and finish about $3$\,s sooner end to end ($14.3$\,s vs.\ $17.3$\,s), so the speedup over the unpruned model is shared by all pruning methods.

\begin{table}[t]
\centering
\footnotesize
\setlength{\tabcolsep}{3.2pt}
\caption{End-to-end efficiency on Video-MME ($N{=}100$, $20\%$ budget).
Query-based methods additionally pay a shared $3.06$\,s decomposition cost (included in
E2E). TFLOPs and generation time are nearly identical across pruned
methods because they share the same retained-token count.}
\label{tab:addl-eff}
\begin{tabular}{lrrrr}
\toprule
Method & E2E\,(s) & Gen\,(s) & Mem\,(GB) & TFLOPs \\
\midrule
\textsc{NoPrune}   & 21.40 & 8.03 & 40.72 & 1207.8 \\
\midrule
\textsc{PruneVid}  & 14.28 & 0.90 & 27.03 & 264.6 \\
\textsc{VisionZip} & 14.26 & 0.89 & 27.03 & 264.6 \\
\textsc{DyCoke}    & 14.28 & 0.91 & 27.03 & 264.6 \\
\textsc{HoliTom}   & 14.28 & 0.90 & 27.03 & 264.6 \\
\textsc{Structured} & 17.32 & 0.89 & 27.03 & 264.6 \\
\textsc{QuoTA}     & 17.34 & 0.90 & 27.03 & 264.6 \\
\textsc{Pareto}    & 17.42 & 0.90 & 27.03 & 264.6 \\
\bottomrule
\end{tabular}
\end{table}

\begin{table}[t]
\centering
\scriptsize
\setlength{\tabcolsep}{2.2pt}
\caption{Decomposition-overhead reduction (Video-MME, $N{=}100$, $20\%$ budget, \textsc{Pareto}); Acc in \% }
\label{tab:addl-cot}
\begin{subtable}[t]{0.40\linewidth}
\centering
\caption{Decomposer size.}
\begin{tabular}{lrr}
\toprule
Decomposer & Acc & Decomp.\,(s) \\
\midrule
Qwen3-0.6B & 61.0 & 3.20 \\
Qwen3-1.7B & 60.0 & 3.04 \\
Qwen3-4B   & 62.0 & 4.02 \\
VLM (8B)   & 61.0 & 5.44 \\
\bottomrule
\end{tabular}
\end{subtable}
\hfill
\begin{subtable}[t]{0.22\linewidth}
\centering
\caption{Batching.}
\begin{tabular}{lr}
\toprule
Batch & s/query \\
\midrule
1  & 3.23 \\
4  & 1.18 \\
8  & 0.77 \\
16 & 0.69 \\
\bottomrule
\end{tabular}
\end{subtable}
\hfill
\begin{subtable}[t]{0.30\linewidth}
\centering
\caption{Token budget.}
\begin{tabular}{lrr}
\toprule
\texttt{mnt} & Acc & Decomp.\,(s) \\
\midrule
512 & 61.0 & 3.03 \\
256 & 61.0 & 2.86 \\
128 & 62.0 & 1.82 \\
\bottomrule
\end{tabular}
\end{subtable}
\end{table}


\section{Conclusion, Limitations, and Future Work}
\label{sec:conclusion}


We studied when and how query information helps training-free, pre-LLM video token pruning. Raw-query alignment is unreliable, while structured decomposition provides a more robust query signal; nevertheless, strong query-agnostic coverage methods remain competitive. Our results suggest a simple design principle: favor strong visual coverage at moderate budgets, introduce structured relevance when greater selectivity is needed for tight budgets or sparse evidence, and combine relevance and coverage through Pareto selection when both matter. Across these regimes, token pruning substantially reduces prefill compute and memory relative to full-token inference.

Our method relies on the decomposer producing anchors that capture the needed evidence and on the visual encoder aligning text anchors with visual evidence; both are prone to hallucination. Combining structured query selection with redundancy-based pruning at more aggressive compression ratios; granular task and video length specific analysis; instruction-tuning compact decomposers to reduce anchor hallucination and policy errors; combining ante-hoc query analysis with intra-LLM or KV-cache pruning; and extending to other multimodal and omnimodal generation and reasoning settings and variable-length visual tokenization are avenues for future work.

\noindent \textbf{AI Usage.} We used AI tools for coding, figure polishing, and language editing; the authors verified all technical content, results, and claims.
{
    \small
    \bibliographystyle{ieeenat_fullname}
    \bibliography{references,references1}
}
\clearpage
\appendix
\section{Testbed and Experimental Setup}
\label{app:testbed}
\label{sec:unified-methods}
\label{app:setup}
\label{sec:unified}

\subsection{Re-implementation Principles and Shared Signals}
\label{app:a-principles}
To study the broad design space on equal footing, we re-implement the \emph{core selection principle} of a diverse set of prior methods on the \emph{same} frozen Qwen3-VL visual tokens, under an identical video configuration, budget rule, and decoding protocol, so that methods differ \emph{only} in which tokens they keep. Components that merge tokens or run inside the LLM are omitted. These are training-free re-implementations inspired by each method's core principle on the shared Qwen3-VL feature space. The blend weights and thresholds below (e.g., the $0.6/0.4$ split in \textsc{VisionZip} and $\tau{=}0.98$) are fixed, hand-set values chosen from prior work and common practice.  

Methods are organized along the query-usage axis of \autoref{sec:taxonomy}. Within the query-based group, some carry their \emph{own} query signal (e.g., a learned selector), while \emph{host} algorithms receive our structured-query prior~$r$ so that we can test its transferability. 

\paragraph{Shared per-token signals.}
From the token embeddings $V\in\mathbb{R}^{N\times D}$ we precompute three min--max-normalized maps per video:
\begin{itemize}[leftmargin=1.2em,itemsep=1pt,topsep=2pt]
\item \textbf{Magnitude} $m_i = \lVert V_i\rVert_2$ (visual ``energy''), the intrinsic saliency $S_{\text{init}}$.
\item \textbf{Uniqueness} $u_i = (1-s_i)\,\mathbb{1}[s_i<\tau]$, where $s_i=\max(s^{\text{tmp}}_i, s^{\text{spa}}_i)$ is the larger of the cosine similarity to the co-located token in the previous frame and the max cosine to its four spatial neighbours; $\tau{=}0.98$. This is the redundancy diversity score $S_{\text{red}}$.
\item \textbf{Anchor relevance} $r_i$: for each query anchor (entity/action with its correlates) we take the clamped cosine between $V_i$ and the pooled anchor text embedding, add a correlate ``boost,'' and sum over anchors ($r_i\!\equiv\!0$ under \textsc{Global} policy or when no decomposition is available). This is $S_{\text{sem}}$, \emph{our} query signal.
\end{itemize}
The \emph{combined} query-conditioned score is $c_i = \mathrm{norm}(m_i + r_i)$, i.e.\ the fused importance $S$. Methods that use $r$ (equivalently $c$) consume our structured-query prior; methods that use only $m$ and $u$ do not.

\subsection{Backbone and Video Configuration}
\label{sec:unified-config}

All experiments share a single backbone, video configuration, budget rule, and decoding protocol, so that methods differ only in which tokens they keep. Here we specify that common protocol.

All experiments use \textbf{Qwen3-VL-8B-Instruct} as the primary Video-LLM backbone, with additional cross-architecture validation on \textbf{InternVL3-8B}~\cite{chen2024internvl}. Our framework operates as a training-free plug-in positioned between the visual encoder and language model: given a user query, the system performs query decomposition (for query-based methods), computes token importance, selects tokens, and forwards them to the Video-LLM for generation. No model weights are modified.

\paragraph{Video Processing Configuration.}
We configure Qwen3-VL with a volume-based adaptive resolution scheme: the total visual volume is capped at $200{,}000 \times 32^2$ tokens, with per-frame limits of \texttt{max\_pixels}$= 768 \times 32^2 = 786{,}432$ (up to 768 tokens per frame) and \texttt{min\_pixels}$= 128 \times 32^2 = 131{,}072$ (at least 128 tokens per frame). The native patch size is $16 \times 16$ pixels. Frames are sampled at 1~FPS with a maximum of 2048 frames and a minimum of 16 frames per video. Tokens are ordered $[t, h_g, w_g]$ with post-pooling grid $h_g{=}h/2$, $w_g{=}w/2$; $N{=}t\,h_g w_g$.

\paragraph{InternVL3-8B Configuration.}
For cross-architecture validation, we use \texttt{OpenGVLab/InternVL3-8B-hf} in bfloat16 precision with automatic device mapping. Videos are sampled at approximately 1~FPS, capped at a maximum of 32 frames. Each frame produces 256 visual tokens on a $16 \times 16$ grid. The same query-decomposition pipeline and token selection module are applied without architecture-specific tuning.

\paragraph{Token Budget and Granularity.}
The token retention budget is specified as a percentage of the total visual tokens produced by the vision encoder after spatial pooling. To ensure efficient GPU utilization with tensor-core operations, the number of retained tokens is rounded to the nearest multiple of 32. Formally we use the budgeted top-$k$ rule
$k = \min\!\big(N,\ \max(32,\ \mathrm{round}(\lfloor N \rho \rfloor/32)\cdot 32)\big)$,
where $N$ is the number of visual tokens and $\rho$ the retention budget.

\paragraph{Query Decomposition Stage.}
Query decomposition follows \autoref{sec:methodology}: a text-only, 7-shot procedure emitting JSON, run with Qwen3 thinking mode disabled in all experiments, so the decomposer is a plain instruction-following step. Its capacity is itself a variable: we report results with a text-only \textbf{Qwen3-0.6B} decomposer (the default for \autoref{sec:rq3}--\ref{sec:anatomy}) and with the \textbf{Qwen3-VL-8B} model in text-only mode (\autoref{sec:analysis-core}); the decomposer is named in each table caption. Unless otherwise noted, the answer is decoded greedily (\texttt{max\_new\_tokens}$=10$ for the multiple-choice answer) with seed $42$, and decomposition generation uses \texttt{max\_new\_tokens}$=512$.

\paragraph{Datasets and metric.}
We evaluate on three video understanding benchmarks. \textbf{MVBench} contains diverse tasks requiring object recognition, motion perception, action recognition, temporal ordering, and visual reasoning over short and medium-length videos (2996 samples total; some analyses use the 2706 samples whose decomposition/answer outputs parsed successfully). \textbf{LongVideoBench} focuses on long-context understanding with extended temporal dependencies (1337 samples), a particularly stringent setting for pruning. \textbf{Video-MME} is a large-scale benchmark covering a broad spectrum of perception and reasoning tasks (900 videos, 300 each of short, medium, and long duration, with 2700 QA pairs, 900 per duration; evaluated without subtitles). Following prior work, we report \textbf{accuracy (\%)} and vary the token retention budget; higher pruning ratios correspond to greater computational savings.

\section{Level 1: Query-Agnostic Methods}
\label{sec:cat-agnostic}

These methods select tokens from intrinsic visual properties alone; the question never enters. Each is reduced to its pre-LLM selection rule on the shared features. Following the design of our testbed, we do not use components that merge or rewrite tokens, or that run inside the LLM, because they either change the token content or fall outside a study of pre-LLM selection; this applies equally to every method, and is noted per method below.

\begin{itemize}[leftmargin=1.2em,itemsep=2pt,topsep=2pt]
\item \textbf{\textsc{PruneVid}}~\cite{huang2025prunevid}: top-$k$ by uniqueness $u$. \emph{Not used:} clustering/merging (it rewrites tokens rather than selecting them) and the LLM attention read-out (it runs inside the LLM).
\item \textbf{\textsc{VisionZip}}~\cite{yang2025visionzip}: keep $0.6k$ ``dominant'' tokens by magnitude $m$, then $0.4k$ ``contextual'' tokens by uniqueness $u$ among the remainder. \emph{Adapted:} magnitude stands in for encoder-attention scoring so that selection stays on the shared token features. \emph{Not used:} merging of the remaining tokens (it rewrites tokens).
\item \textbf{\textsc{DyCoke}}~\cite{tao2025dycoke}: top-$k$ by \emph{temporal} uniqueness $1-\cos(V_i, V_i^{\text{prev-frame}})$, with a $10^{-3}m$ tie-break. \emph{Not used:} temporal merging (it rewrites tokens) and decoding-time KV-cache pruning (it runs inside the LLM).
\item \textbf{\textsc{HoliTom}}~\cite{shao2025holitom}: enforce an even per-frame quota $\lfloor k/t\rfloor$ selected by $u$, then globally fill by $u$. \emph{Not used:} temporal segmentation and spatio-temporal merging (token merging) and inner-LLM merging (it runs inside the LLM).
\item \textbf{\textsc{TopV}}~\cite{yang2025topv}: transport-based ``prune-once'' selection, re-implemented as a Sinkhorn-style optimal-transport assignment over the magnitude/uniqueness feature map. \emph{Adapted:} the assignment runs ante-hoc on the shared features instead of inside the LLM (listed as intra-LLM in \autoref{sec:related-intrallm}).
\item \textbf{\textsc{IVC-Prune}}~\cite{sun2026ivc}: position-aware selection that protects implicit visual coordinates within the RoPE grid while pruning by intrinsic saliency.
\item \textbf{\textsc{GAP}}~\cite{chien2025grounding}: grounding-aware, position-preserving feature selection (``feature-norm'' variant) that freezes the coordinate grid during pruning.
\item \textbf{\textsc{Representation-Shift}}~\cite{choi2025representation}: prune by Euclidean shift of token embeddings in representation space, a purely intrinsic embedding-space criterion. \emph{Adapted:} we use the shared token embeddings in place of the model's internal embedding states, which live inside the LLM (listed as intra-LLM in \autoref{sec:related-intrallm}).
\end{itemize}
\section{Level 2: Shallow Query Conditioning (Methods with Their Own Query Signal)}
\label{sec:cat-querybased}

These methods let the question influence selection. Each is reduced to an ante-hoc scorer on the shared features. Where a method's original signal lives inside the LLM, we use an ante-hoc proxy in its place, because our study selects tokens before the LLM; this applies equally to every method below and is noted per method.

\paragraph{Own query signal (shallow).}
\begin{itemize}[leftmargin=1.2em,itemsep=2pt,topsep=2pt]
\item \textbf{\textsc{FastV}} (raw-query)~\cite{fastv2024}: top-$k$ by the clamped, normalized cosine between each token and the pooled embedding of the \emph{raw question text}, with a $10^{-3}m$ tie-break. \emph{Adapted:} the original prunes inside the LLM after layer~2 using attention, so we use an ante-hoc raw-query$\to$token cosine instead.
\item \textbf{\textsc{FastV-Attn}}~\cite{ye2025atp}: the attention-style variant, using a cross-modal attention read-out between question and tokens as the relevance signal.
\item \textbf{\textsc{FlashVLM}}~\cite{cai2025flashvlm}: cross-modal cosine filtering between the question embedding and visual tokens.
\item \textbf{\textsc{FlexSelect}}~\cite{zhang2025flexselect}: a learned selector mimicking cross-modal attention. \emph{Adapted:} re-implemented as its query-relevance scoring rule on the shared features.
\item \textbf{\textsc{ReTaKe}}~\cite{wang2024retake}: keyframe-based temporal selection (the ``DP-Select'' principle), isolating query-relevant keyframes before pruning.
\end{itemize}
\section{Level 3: Structured Query Pruning}
\label{app:method}
This appendix gives the full details of Structured Query Pruning (\autoref{sec:methodology}), including the decomposition prompt.

\begin{figure*}[t]
    \centering
    \includegraphics[width=0.9\textwidth]{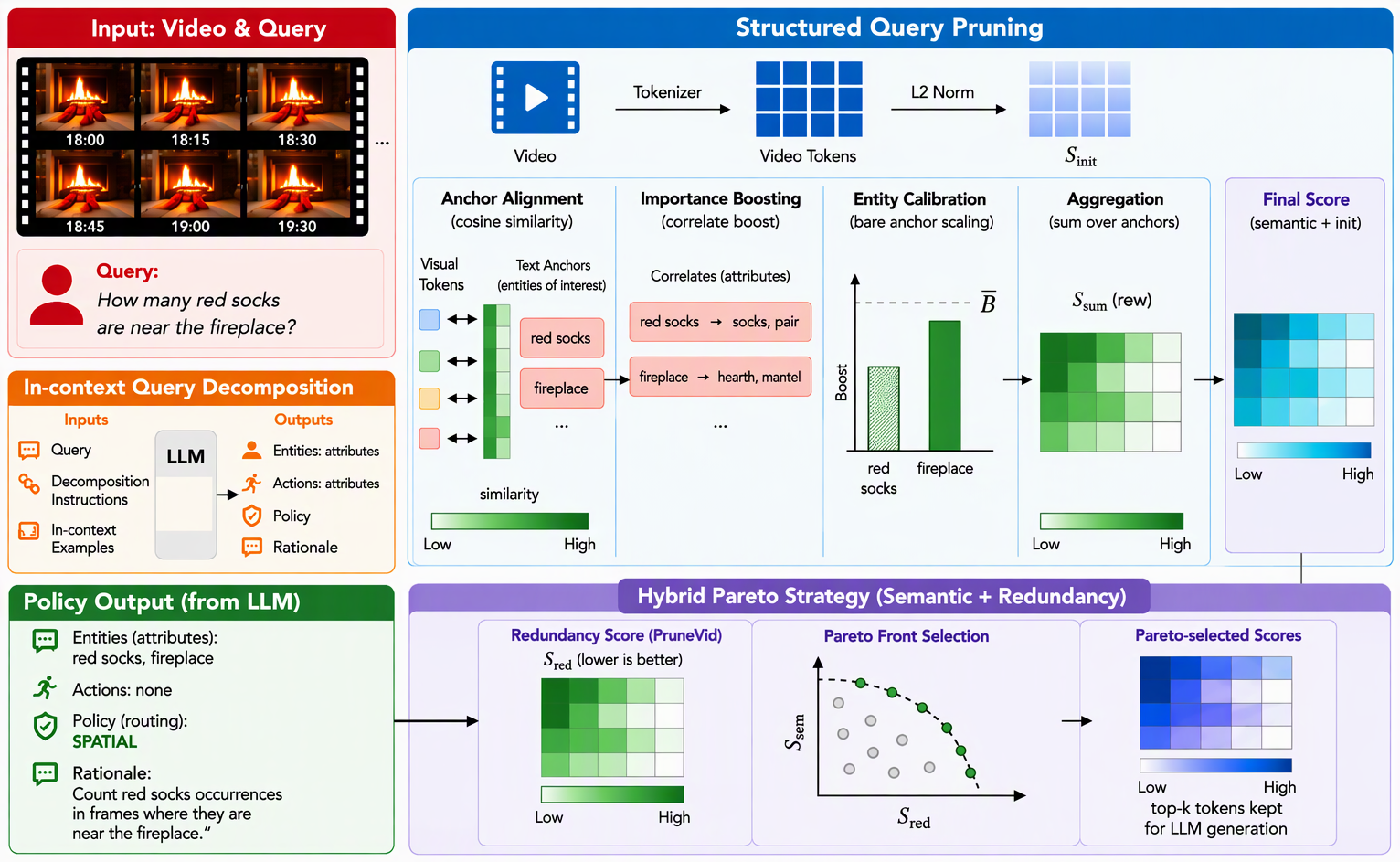}
    \caption{\textbf{Overview of Structured Query Pruning.} Given a user query, the system performs (1)~ante-hoc query decomposition to extract semantic anchors (entities, actions) and a task policy, then (2)~computes a fused importance score combining intrinsic visual saliency with query-derived semantic relevance. Tokens are ranked and the top-$k$ are retained for LLM generation. The hybrid Pareto strategy jointly optimizes semantic relevance and visual diversity. }
    \label{fig:overview}
\end{figure*}

\subsection{Query Decomposition}
\label{app:b-decomp}

Decomposition is text-only: video frames are withheld to minimize overhead and avoid visual hallucination. Given the query, an LLM generates a structured JSON object (few-shot, in-context; Qwen3 thinking mode disabled) with the following components.

\subsubsection{Policy Selection}
The model first classifies the task into one of four distinct strategies to determine the ``flavor'' of the required visual analysis:
\begin{itemize}
    \item \textbf{SPATIAL}: Prioritizes static visual cues, localization, and attribute identification (e.g., ``What color is the object?'').
    \item \textbf{TEMPORAL}: Focuses on motion characteristics and kinematic sequences (e.g., ``Is the person running?'').
    \item \textbf{BALANCED}: Requires an integrated analysis of both spatial attributes and temporal dynamics (e.g., ``What color shirt is the running person wearing?'').
    \item \textbf{GLOBAL}: Targets coarse-grained semantic understanding or video summarization (e.g., ``What is the genre of this video?'').
\end{itemize}

\subsubsection{Semantic Anchor Identification}
Once the policy is established, the model identifies specific visual anchors to search for within the video. We prompt the LLM to extract the following fields:
\begin{itemize}
    \item \textbf{Entities:} Bare-noun visual anchors (e.g., ``socks'') representing the primary objects of interest.
    \item \textbf{Entity Correlates:} Disambiguating attributes (e.g., \{``socks'': [``red'']\}) used only when the bare noun requires further specification based on the query.
    \item \textbf{Actions:} Bare-verb kinematic signals (e.g., ``jumping''). To ensure precision, we explicitly prohibit the use of generic meta-verbs such as ``showing'' or ``talking.''
    \item \textbf{Action Correlates:} Specific physical modifiers (e.g., \{``jumping'': [``upward'']\}) that characterize the motion.
\end{itemize}

To ensure the model remains grounded in the user's intent, the LLM is strictly instructed to derive these anchors from the query without introducing extrinsic information.

\subsubsection{Implementation and Parsing}
The LLM is prompted with diverse in-context examples of policies and their anchors, and the JSON block is extracted by brace-counting. The result is a dictionary mapping:
\begin{equation}
    \{ \text{bare\_noun} : [\text{correlates}] \}
\end{equation}
If no correlates are identified, the entity or action maps to itself, creating a clean, structured representation for the subsequent scoring phase.

\subsubsection{Prompt Template}
\label{sec:cot_prompt}

Below is a representative example of our in-context learning prompt for query decomposition. The LLM receives 7 diverse examples spanning all four policies before processing each test query.

\begin{small}
\begin{verbatim}
System: You are a video analysis planner.
Given a question about a video, decompose 
it into structured search targets WITHOUT 
seeing the video.

Output a JSON with:
- "policy": SPATIAL/TEMPORAL/BALANCED/GLOBAL
- "entities": list of bare-noun targets
- "entity_correlates": {entity: [attrs]}
- "actions": list of bare-verb targets  
- "action_correlates": {action: [mods]}
- "rationale": one-line justification

Rules:
- Only extract from the query text
- No generic verbs (show, display, appear)
- Correlates only when needed
- Map to self if no correlates exist
\end{verbatim}
\end{small}

\subsection{Policy-Driven Importance Scoring}
\label{app:b-scoring}

Given the query anchors, we compute a per-token importance score that integrates intrinsic visual saliency with query-derived semantic relevance, governed by the task policy.

\subsubsection{Intrinsic Visual Saliency ($S_{\text{init}}$)}
As a baseline, we compute the initial saliency scores, $S_{\text{init}}$, using the per-token $L_2$ norms of the visual embeddings $\mathbf{V}$. These norms are normalized to the range $[0, 1]$ per sample. This step captures the \textit{natural saliency} of the video, ensuring that visually dense regions—regardless of the specific query—are preserved as a foundational context for the model.

\subsubsection{Query-Derived Semantic Mapping ($S_{\text{sem}}$)}
To align visual tokens with the semantic objectives identified in the decomposition phase, we compute a semantic mapping score. Let $\mathbf{v}_i$ be the embedding of the $i$-th visual token and $\mathbf{t}_a$ be the text embedding of a semantic anchor (entity or action). The semantic score is derived as follows:

\begin{itemize}
    \item \textbf{Anchor Alignment:} We compute the cosine similarity between every visual token and the target text embedding. Negative similarities are clamped to zero to ensure only positive correlations contribute to retention.
    \item \textbf{Importance Boosting:} For each anchor, we utilize its associated correlates to refine the score. If an entity has correlates (e.g., ``red socks''), the system computes the average similarity between the token and the correlate embeddings to generate a \textit{boost} signal.
    \item \textbf{Entity Calibration:} To prevent ``bare'' anchors (those without correlates) from being statistically overshadowed by attributed anchors, we calibrate them using the attributed anchors' behaviour. For each attributed anchor we record the per-token ratio of its correlate boost to its target similarity; we then average these ratios across all attributed anchors in the query. A bare anchor's boost is synthesized as its own target similarity multiplied by this mean ratio, so that bare anchors receive a boost proportional to their relevance rather than being dropped.
    \item \textbf{Aggregation:} The total semantic importance is the summation of anchor similarities and their respective boosts:
    \begin{equation}
        S_{\text{sem}} = \sum (\text{similarity}_{\text{anchor}} + \text{boost}_{\text{correlate}})
    \end{equation}
\end{itemize}

\subsubsection{Final Score Fusion}
The final importance score $S$ is the sum of the intrinsic saliency and the semantic map, re-normalized to $[0,1]$ per sample:
\begin{equation}
    S = \mathrm{norm}\!\left(S_{\text{init}} + S_{\text{sem}}\right)
\end{equation}
Under the \texttt{GLOBAL} policy, $S_{\text{sem}}\!\equiv\!0$ and $S$ reduces to $S_{\text{init}}$. Visual tokens are ranked according to $S$, and the top-$k$ tokens are retained for the generation phase. This fused score $S$ (and \emph{not} the semantic map $S_{\text{sem}}$ alone) is what we refer to as the ``Structured-Query'' selector in the experiments.

\subsection{The Structured Selector}
\label{app:b-structured}
The \textsc{Structured} selector keeps the top-$k$ tokens by the combined importance $c=\mathrm{norm}(m+r)$ (\autoref{sec:importance_scoring}); it uses the structured-query prior directly.

\subsection{Host Algorithms (Consume the Prior $r$)}
\label{app:c-hosts}

Each host keeps its own selection philosophy and receives $r$ in its native ranking criterion. We adapt only what is needed to run before the LLM: components that depend on an additional LVLM pass or on LLM layers are replaced by token-level equivalents on the shared features, or not used, as noted per method.

\begin{itemize}[leftmargin=1.2em,itemsep=2pt,topsep=2pt]
\item \textbf{\textsc{QuoTA}}~\cite{luo2026quota}: quality $q=\mathrm{norm}(0.6m+0.4r)$; an adaptive threshold sized to $\sim\!1.5k$, a guaranteed $\lfloor k/(2t)\rfloor$ tokens per frame for coverage, filled to $k$ by $q$. \emph{Adapted:} the original's LVLM pass for frame-level relevance is replaced by a token-level quality score from $m{+}r$, since the pass requires an additional LVLM forward. The without-prior variant ($w/o\ r$) runs the same allocator on $r\!\equiv\!0$.
\item \textbf{\textsc{AdaReTaKe}}~\cite{wang2025adaretake}: start from $\tfrac12 c + \tfrac12 u$, then three refinement rounds that boost under-covered frames/cells (gap $\times\,c$, margin $0.8^{\text{round}}$) and re-select top-$k$. \emph{Not used:} allocation across LLM layers and per-layer KV budgeting, which run inside the LLM.
\item \textbf{\textsc{DivPrune}}~\cite{alvar2025divprune}: greedy Max--Min diversity---seed with the top-$c$ token, then add the token maximizing $c_i + 0.4(1-\max_{j\in\text{sel}}\cos(V_i,V_j))$ until $k$ are chosen. This follows the published greedy MMDP selection, with $c$ as the per-token value term.
\item \textbf{\textsc{Tango}}~\cite{yin2026tango}: temporally stable token importance. Let $s_{t,p}$ be the fused importance $c$ of the token at frame $t$ and spatial position $p$, so $r$ enters through $s$. Tango smooths $s$ over a temporal window, $\tilde{s}_{t,p}=0.7\,s_{t,p}+0.3\,\mathrm{Avg}_{\tau\in\mathcal{N}(t)}\,s_{\tau,p}$, computes a temporal-consistency score from the variance of importance across frames, $\kappa_p=1/(\mathrm{Var}_t(s_{t,p})+0.1)$, and scores tokens by $q_{t,p}=\tilde{s}_{t,p}+0.3\,\kappa_p$. The video is divided into four temporal segments with about $k/4$ tokens selected from each, and any unallocated budget is filled by the highest-scoring remaining tokens. Tango therefore favors tokens whose importance stays stable over time while keeping coverage balanced across the video. \emph{Inspired by the principle of the cited work: a temporal-consistency selector.}
\item \textbf{\textsc{SemanticEvidence}}~\cite{li2026keeping}: evidence $=0.6r+0.4u$ accumulated along per-position temporal trajectories (threshold $0.5$); select the highest-scoring evidence chains up to $k$. \emph{Inspired by the principle of the cited work: evidence chains along temporal trajectories.}
\end{itemize}

\section{Level 4: Holistic Selection}
\label{sec:cat-ours}

Our holistic selectors combine the structured-query prior with query-agnostic redundancy (\autoref{sec:holistic}):
\begin{itemize}[leftmargin=1.2em,itemsep=2pt,topsep=2pt]
\item \textbf{\textsc{Pareto}} (holistic): Multi-Round Pareto Layering on $(u,c)$ (\autoref{sec:holistic})---sort by $u$, iteratively extract the Pareto frontier of $(u,c)$, and fill the budget by the product $u\!\cdot\!c$. This is our hybrid of query-agnostic redundancy and structured-query relevance.
\item \textbf{\textsc{Nested}} (holistic): a sequential variant that first filters to a $2k$ pool by uniqueness $u$, then re-ranks that pool by $c$.
\end{itemize}

\subsection{Hybrid Selection: Reconciling Redundancy and Query Relevance through Pareto Selection}
\label{app:b-pareto}

Redundancy-based pruning (e.g., \textit{PruneVid}) removes visual overlap but ignores the user's semantic goals; query-based pruning ensures relevance but may retain redundant tokens that share an anchor. We therefore propose a hybrid, based on multi-objective optimization, for the \texttt{SPATIAL}, \texttt{TEMPORAL}, and \texttt{BALANCED} policies; \texttt{GLOBAL} defaults to redundancy-based sampling. This hybrid is our instance of a \emph{holistic} design in the query-usage spectrum.

\subsubsection{Multi-Round Pareto Layering}

To treat token selection as a formal multi-objective optimization problem, we aim to maximize two potentially conflicting objectives: the redundancy-based diversity score $S_{\text{red}}$ (the normalized PruneVid uniqueness) and the fused query-conditioned importance $S = \mathrm{norm}(S_{\text{init}} + S_{\text{sem}})$ defined above. We identify the optimal token set by iteratively extracting the \textbf{Pareto Frontier}. 

\paragraph{Pareto Optimality and Frontier Extraction.}
We implement frontier extraction efficiently by first sorting all remaining tokens by $S_{\text{red}}$ in descending order. Scanning in this order, a token $i$ is placed on the current frontier if its fused importance $S_i$ strictly exceeds the running cumulative maximum of $S$ over all previously scanned tokens (those with higher $S_{\text{red}}$), using a numerical margin of $10^{-6}$. Equivalently, a token is \textit{Pareto-optimal} if no other token has both higher $S_{\text{red}}$ and higher $S$; the sorted cumulative-maximum scan recovers exactly this non-dominated set.

\paragraph{Iterative Selection and Product-Score Tie-Breaking.}
The selection process follows an iterative layering approach:
\begin{enumerate}
\item \textbf{Frontier Identification:} We compute the Pareto frontier of the remaining token pool as above.
\item \textbf{Quota Comparison:} If the number of tokens on the frontier is less than or equal to the remaining budget $K_{\text{rem}}$, the entire frontier is accepted and removed from the pool; the next round operates on the tokens left behind.
\item \textbf{Product-Score Tie-Breaker:} If a frontier exceeds the remaining budget, we rank its tokens by the \textbf{product} of the two normalized scores and keep the \emph{highest}-scoring ones until the budget is met:\begin{equation}P(i) = S_{\text{red},i}\,\cdot\,S_{i}.\end{equation}
The same product $P$ is used to fill any residual budget after the final frontier. 
\end{enumerate}

\section{Extended Results on the Three Benchmarks}
\label{app:results}
\label{sec:analysis-core}

These per-benchmark results extend RQ1 (\autoref{sec:rq1}). We begin with a per-benchmark comparison using the \textbf{Qwen3-VL-8B} model, the same model in text-only mode is used as the decomposer. We organize the evaluated methods into three categories: the \textbf{Redundancy-Aware} baseline (\textsc{PruneVid} uniqueness, query-agnostic); the \textbf{Structured-Query} selector (structured query); and two \textbf{Holistic} strategies, \textbf{Nested} (sequential coarse-to-fine: PruneVid filtering to a $2K$ pool, then re-ranking by fused $S$) and \textbf{Pareto} (joint multi-objective, our proposed holistic method).

Rather than focusing solely on absolute accuracy, our analysis asks \emph{when} semantic relevance, visual diversity, or their combination becomes the dominant factor.

\subsection{Video-MME: Redundancy Is Competitive at Mild Compression}
\label{app:e-videomme}

\begin{table}[t]
\centering
\caption{\textbf{Video-MME accuracy (\%)}, Qwen3-VL-8B. Best result per row in \textbf{bold}.}
\label{tab:videomme_results}
\resizebox{\columnwidth}{!}{ 
\begin{tabular}{c|c|c|>{\columncolor{gray!12}}c|>{\columncolor{gray!12}}c}
\toprule

& \multicolumn{1}{c|}{\textbf{Redundancy-Aware}}
& \multicolumn{1}{c|}{\textbf{Structured-Query}}
& \multicolumn{2}{c}{\textbf{Holistic (Redundancy + Query Analysis)}} \\

\textbf{Budget (\%)}
& \textbf{PruneVid}
& \textbf{Structured}
& \textbf{Pareto}
& \textbf{Nested} \\

\midrule

10 & 62.1 & 60.6 & \textbf{64.3} & 62.3 \\
20 & 64.4 & 61.8 & \textbf{65.0} & 63.7 \\
30 & 65.0 & 63.1 & \textbf{65.8} & 64.5 \\
40 & 65.4 & 64.1 & \textbf{66.2} & 65.2 \\
50 & 66.1 & 64.1 & \textbf{66.2} & 64.7 \\
60 & \textbf{66.1} & 65.2 & 65.7 & 65.1 \\
70 & 66.0 & 66.1 & 65.8 & \textbf{66.5} \\
80 & 66.0 & 66.8 & 66.3 & \textbf{66.9} \\
90 & \textbf{66.3} & 65.8 & 66.0 & 65.8 \\
\midrule 
100 (No pruning) & \multicolumn{4}{c}{71.4} \\
\bottomrule
\end{tabular}
}
\end{table}

Video-MME evaluates broad video understanding including scene interpretation, activity recognition, and multimodal reasoning (Table~\ref{tab:videomme_results}). Structured-query pruning alone is \emph{below} redundancy-aware pruning at every budget from 10\% to 60\%, suggesting that semantic relevance alone is insufficient for holistic video understanding. The strongest results come from the holistic hybrid: Pareto achieves the best accuracy across most low-budget settings (10--50\%), improving on either relevance or redundancy alone, although a no-query magnitude-plus-redundancy reference performs comparably (64.2\% and 65.3\% at 10\% and 20\% retention). At 10\% retention, Pareto achieves 64.3\% versus 62.1\% for PruneVid and 60.6\% for pure structured-query selection.

\subsection{MVBench: Structured Queries Beat the Redundancy Baseline}
\label{app:e-mvbench}

\begin{table}[t]
\centering
\caption{\textbf{MVBench accuracy (\%)}, Qwen3-VL-8B .  Best per row in \textbf{bold}.}
\label{tab:mvbench_results}
\resizebox{\columnwidth}{!}{ 
\begin{tabular}{c|c|c|>{\columncolor{gray!12}}c|>{\columncolor{gray!12}}c}
\toprule

& \textbf{Redundancy}
& \textbf{Structured query}
& \multicolumn{2}{c}{\textbf{Holistic}} \\

\cline{2-5}

\textbf{Budget (\%)}
& \textbf{PruneVid}
& \textbf{Structured}
& \textbf{Pareto}
& \textbf{Nested} \\

\midrule

10 & 48.5 & \textbf{52.6} & 51.8 & 51.0 \\
15 & 50.6 & \textbf{55.2} & 54.8 & 53.1 \\
20 & 52.5 & \textbf{57.3} & 56.0 & 55.1 \\
25 & 53.6 & \textbf{57.4} & 56.3 & 56.2 \\
30 & 54.9 & \textbf{58.3} & 58.2 & 57.3 \\
40 & 56.7 & \textbf{59.4} & 58.7 & 58.7 \\
50 & 57.8 & \textbf{60.6} & 59.7 & \textbf{60.6} \\
\midrule 
100 (No pruning) & \multicolumn{4}{c}{60.6} \\

\bottomrule
\end{tabular}
}
\end{table}

Unlike Video-MME, structured-query token selection consistently outperforms redundancy-based pruning across all budgets on MVBench, with improvements of 2--5 percentage points (Table~\ref{tab:mvbench_results}). Large gains occur under aggressive pruning (4.1 points at 10\% retention, and the largest margin of 4.8 points at 20\%). This is consistent with the nature of MVBench tasks, which require identifying specific objects, localized actions, or short temporal events where preserving semantically relevant evidence matters more than maintaining broad visual coverage. This comparison is against the redundancy baseline specifically; against the \emph{full} field of competing pruning principles the structured selector is mid-field rather than dominant.

\subsection{LongVideoBench: Largest Gains at the Tightest Budget}
\label{app:e-lvb}

\begin{table}[t]
\centering
\caption{\textbf{LongVideoBench accuracy (\%)}.  Full-token baseline: 53.4\%. Best per row in \textbf{bold}.}
\label{tab:longvideo_results}
\resizebox{\columnwidth}{!}{ 
\begin{tabular}{c|c|c|>{\columncolor{gray!12}}c|>{\columncolor{gray!12}}c}
\toprule

& \textbf{Redundancy}
& \textbf{Structured query}
& \multicolumn{2}{c}{\textbf{Holistic}} \\

\textbf{Budget (\%)}
& \textbf{PruneVid}
& \textbf{Structured}
& \textbf{Pareto}
& \textbf{Nested} \\

\midrule

10 & 50.6 & \textbf{55.5} & 53.6 & 53.5 \\
15 & 53.6 & \textbf{56.9} & 54.5 & 54.5 \\
20 & 54.4 & \textbf{56.5} & 56.4 & 55.4 \\
25 & 54.8 & \textbf{56.5} & 56.3 & 55.3 \\
30 & 55.2 & 56.5 & \textbf{56.7} & 55.9 \\
40 & 54.5 & \textbf{56.4} & 55.7 & 55.4 \\
50 & 53.7 & 55.3 & 55.2 & \textbf{55.7} \\
\midrule
100 (No pruning) & \multicolumn{4}{c}{53.4} \\

\bottomrule
\end{tabular}
}
\end{table} 

Structured-query selection achieves the highest accuracy across nearly all budgets on LongVideoBench, with the largest gain at 10\% retention (+4.9 points over PruneVid; Table~\ref{tab:longvideo_results}). Holistic approaches perform competitively but generally do not surpass pure structured-query selection here, suggesting that for long-context tasks semantic relevance dominates over visual diversity. A plausible reason is that critical evidence appears only briefly amid large amounts of irrelevant content, so query anchors may turn pruning into targeted evidence retrieval, preserving sparse observations regardless of temporal location. 

\subsection{Cross-Benchmark Synthesis}
\label{app:e-synth}

The effectiveness of pruning strategies is task-dependent. Holistic hybrids are strongest on Video-MME among the categories compared here, structured-query selection consistently beats the redundancy baseline on MVBench, and the trend is most pronounced on LongVideoBench. Overall, redundancy-aware methods determine which information is \textit{unique}, while structured-query methods determine which is \textit{useful}; combining them improves on either alone on Video-MME.

\paragraph{When structured queries help most.} The advantage of structured-query selection is most pronounced (i)~on tasks requiring identification of \emph{specific} visual evidence---on MVBench subtasks involving fine-grained object recognition, temporal ordering, and action localization, structured-query pruning outperforms PruneVid at aggressive compression, while for holistic scene/genre classification redundancy remains competitive (consistent with routing \texttt{GLOBAL} queries to diversity); (ii)~as \emph{video length} grows---on LongVideoBench the structured-query selector reaches 4.9 points over PruneVid at 10\% retention, possibly because longer videos contain proportionally more query-irrelevant content; and (iii)~as \emph{compression severity} increases---at mild compression (70--90\%) all methods are comparable, but below 40\% retention structured-query selection retains more accuracy than the redundancy-only baseline, which degrades faster. 

\end{document}